\documentclass[letterpaper, 10 pt, journal, twoside]{IEEEtran}
\usepackage{tabularx}
\usepackage{booktabs}
\usepackage{cite}
\usepackage{graphicx}
\usepackage{amsmath} % assumes amsmath package installed
\usepackage{amssymb}  % assumes amsmath package installed
\usepackage[colorlinks=true, linkcolor=blue, anchorcolor=blue, citecolor=blue]{hyperref}

\IEEEoverridecommandlockouts                              % This command is only needed if 
\title{
Extending the Speed Limit of Quadrupedal Locomotion via Refined Actuator Modeling and Adaptive Command Scheduling
}

\author{Yucheng Tao$^{1}$, Shaowen Cheng$^{2}$, Guorong Lan$^{3}$, Yanyan Yuan$^{3}$, Yongbin Jin$^{1, 2, *}$, and Hongtao Wang$^{1, 2, *}$ % <-this % stops a space
\thanks{Manuscript received: April, 8, 2026; Revised June, 22, 2026; Accepted July, 27, 2026.}%Use only for final RAL version
\thanks{This paper was recommended for publication by Editor Clement Gosselin upon evaluation of the Associate Editor and Reviewers’ comments.}
\thanks{$^{1}$Authors are with the Center for X-Mechanics and
Institute of Applied Mechanics, Zhejiang University, Hangzhou
310012, China (e-mail: (yctao@zju.edu.cn))}%
\thanks{$^{2}$Authors are with ZJU-Hangzhou Global Scientific and Technological Innovation Center, Hangzhou
311200, China}
\thanks{$^{3}$Authors are with  MirrorMe Robotics Co., Ltd., Hangzhou 311215, China.}
\thanks{$^{*}$Corresponding authors: Yongbin Jin (yongbinjin@zju.edu.cn) and Hongtao Wang (htw@zju.edu.cn).}
\thanks{Digital Object Identifier (DOI): see top of this page.}
}

\begin{document}

\maketitle
% \pagestyle{empty}
% \thispagestyle{empty}
% Comment or remove these lines for final RAL version.

%%%%%%%%%%%%%%%%%%%%%%%%%%%%%%%%%%%%%%%%%%%%%%%%%%%%%%%%%%%%%%%%%%%%%%%%%%%%%%%%
\begin{abstract}
Achieving high-speed locomotion in quadrupedal robots remains highly challenging, as actuators operate near their physical limits and exhibit pronounced nonlinearities.
However, many existing methods neglect actuator nonlinearities and physical constraints during training, leading to a significant sim-to-real gap under highly dynamic motions and limiting achievable performance.
To address this issue, we propose a high-speed locomotion framework that reduces sim-to-real discrepancies and stabilizes learning over a wide command distribution. 
A refined actuator model explicitly captures high-speed voltage coupling and magnetic saturation, enabling a more accurate representation of the torque–speed envelope. In addition, a reinforcement learning framework incorporating a two-stage curriculum and adaptive command scheduling (ACS) ensures stable training. Experiments on the 36.5 kg quadruped BlackPanther2 (BP2) demonstrate speeds of up to $13.2~\mathrm{m/s}$ on a treadmill and $11.65~\mathrm{m/s}$ outdoors, establishing a new state-of-the-art and, to the best of our knowledge, a world record for quadrupedal robot locomotion.
The results further highlight the importance of accurate actuator modeling in preventing non-physical policy exploitation, and show that ACS improves robustness without sacrificing performance.

\end{abstract}

% Keywords appear just beneath the abstract. Use only for final RAL version. 

% \begin{keywords}

% Quadruped robots, high-speed locomotion, reinforcement learning, sim-to-real transfer, actuator modeling, motor dynamics, curriculum learning, ACS

% \end{keywords}
%%%%%%%%%%%%%%%%%%%%%%%%%%%%%%%%%%%%%%%%%%%%%%%%%%%%%%%%%%%%%%%%%%%%%%%%%%%%%%%%
\section{INTRODUCTION}

\IEEEPARstart{S}{ignificant} advances in quadrupedal locomotion have shifted research toward highly dynamic behaviors. Achieving extreme locomotion performance in tasks such as high-speed sprinting \cite{margolis_rapid_2022,shin_reinforcement_2025,ji_concurrent_2022} and agile parkour \cite{zhuang_robot_2023,hoeller_anymal_2024,luo_pie_2024,wang_sf-tim_2024,kim_high-speed_2025,wang_puma_2026,cheng_extreme_2024,he_attention-based_2025} is critical for time-sensitive applications and for expanding operational capabilities in outdoor exploration.

Control paradigms for highly dynamic locomotion have transitioned from classical model-based optimization to data-driven learning. A representative model-based milestone is WildCat \cite{noauthor_introducing_nodate}, which achieved an outdoor speed of approximately 29 km/h. More recently, deep reinforcement learning (DRL) has demonstrated strong capability in handling highly nonlinear dynamics and has significantly improved the agility of electrically actuated quadrupedal robots.
As quadrupedal robots approach their maximum velocities, two primary challenges emerge. First, accurate actuator modeling and effective utilization of actuator capabilities become critical. Under high-speed conditions, motor states approach physical limits \cite{jin_high-speed_2022}, where nonlinear effects—including saturation and torque–speed envelope nonlinearities—dominate system dynamics \cite{chignoli_mit_2021}. Conventional control strategies typically operate within conservative regimes far from the motor operating region (MOR), neglecting torque–speed envelope constraints and thereby limiting performance in extreme regimes. Second, algorithmic stability remains a key challenge. Although DRL shows potential for highly dynamic control, handling a wide command range—from standstill to over $10~\mathrm{m/s}$—often leads to training instability or poor convergence. This issue complicates the trade-off between control performance and robustness.

   % \begin{figure}[t]
   %    \centering
   %    \includegraphics[width=\columnwidth]{gym.png}
   %    \caption{Snapshots of BP2 during high-speed locomotion on an outdoor athletics track are shown at 1/6 s intervals between consecutive frames. The robot achieves an average speed of approximately 10.8 m/s, as measured from high-speed video recorded at 240 fps.}
   %    \label{figure1}
   % \end{figure}

To address these issues, we present a unified framework for safely and stably driving quadrupedal robots toward their physical speed limits. First, we develop a refined actuator model that explicitly captures nonlinearities near physical limits, including $d$-axis voltage coupling and magnetic saturation. These effects are incorporated as torque constraints in simulation, thereby reducing the sim-to-real gap at high speeds.
Second, we introduce a tailored DRL architecture with a two-stage curriculum to progressively expand the feasible motion space. To mitigate command mismatch, an ACS strategy is proposed to decouple robustness from empirical tuning. Instead of imposing fixed acceleration limits, ACS dynamically constrains the command neighborhood relative to the robot's instantaneous velocity, enabling the policy to learn acceleration profiles consistent with hardware constraints.

The proposed framework is deployed on a custom-developed 36.5 kg quadrupedal robot, BP2. Extensive indoor and outdoor experiments demonstrate that the method maintains stability near actuator limits and significantly reduces instability during high-speed locomotion. The main contributions of this work are summarized as follows:
\begin{itemize}
\item An explicit actuator model incorporating $d$-axis voltage coupling and magnetic saturation is proposed, enabling accurate characterization of the torque–speed envelope and reducing modeling discrepancies under high-speed conditions.
\item A training framework integrating a two-stage curriculum with adaptive command scheduling (ACS) is developed, improving stability over a wide command range, preventing policy collapse, and reducing reliance on empirical acceleration tuning.
\item The proposed approach is validated on the BP2 platform, achieving sprinting speeds of $13.2~\mathrm{m/s}$ indoors and $11.65~\mathrm{m/s}$ outdoors, establishing a new state-of-the-art and, to the best of our knowledge, a world record for quadrupedal robot sprinting.
\end{itemize}

\section{RELATED WORK}

\subsection{High-Speed Quadrupedal Locomotion Control}

High-speed locomotion methods for quadruped robots have evolved from early heuristic and physics-based approaches to optimization-based methods, and more recently to data-driven paradigms. Early studies primarily employed Model Predictive Control (MPC) based on simplified Single Rigid Body Dynamics (SRBD) to optimize centroidal dynamics, achieving stable steady-state running \cite{park_high-speed_2017,bledt_mit_2018,kim_highly_2019}. However, as locomotion speed increases, strong impacts and pronounced underactuation degrade the validity of such simplified models. While hydraulic platforms like WildCat leveraged high power-density actuation to achieve early mobility records, their intrinsic response latency and modeling complexity hinder integration with modern high-frequency, lightweight electric actuation systems. In contrast, DRL leverages massively parallel simulation to implicitly capture whole-body nonlinear dynamics \cite{margolis_rapid_2022,hwangbo_learning_2019,shin_reinforcement_2025}, demonstrating the potential to outperform traditional methods for agile locomotion and complex parkour tasks.

\subsection{Sim-to-Real Gap in Motion Control}

The effective deployment of RL controllers on hardware critically depends on bridging the sim-to-real gap. For moderately dynamic tasks, prevailing approaches typically inject observation noise \cite{kumar_rma_2021,margolis_rapid_2022} or apply domain randomization (DR) \cite{peng_sim--real_2018} to improve policy robustness and generalization. However, excessive randomization can limit the achievable performance ceiling, a drawback that becomes more pronounced in highly dynamic locomotion. To prevent policies from exploiting non-physical artifacts in simulation, recent studies incorporate stricter physical constraints. 
For example, \cite{hwangbo_learning_2019} employs actuator networks to learn actuator input--output mappings; \cite{bjelonic_towards_2025} performs system-level frequency sweeps for global identification of the ANYmal platform; and \cite{miller_high-performance_2025} models battery voltage dynamics within the control loop. These approaches improve sim-to-real consistency from different perspectives. However, actuator networks tend to absorb multiple physical effects into a black-box mapping, while system-level identification is high-dimensional and can be affected by coupled robot dynamics, contact, and actuator effects. In contrast, our approach focuses on the actuator operating boundary itself and uses a low-dimensional, physically structured motor model to explicitly capture the high-speed torque--speed envelope.

While modern simulators such as IsaacLab and mjlab \cite{zakka_mjlab_2026} provide built-in motor constraints, and prior works have incorporated actuator operating-region constraints using simplified direct-current (DC) motor models and quasi-direct-drive (QDD) actuator models, including MOR clipping in reinforcement learning \cite{shin_reinforcement_2025,bellegardaRobustQuadrupedJumping2024} and actuator-aware torque-speed constraints in motion planning \cite{chignoli_mit_2021}. However, these approaches do not explicitly focus on the envelope shrinkage caused by d-axis voltage consumption under sustained terminal-speed locomotion. This simplification is acceptable at low-to-moderate speeds, but becomes restrictive under sustained high-speed operation, where cross-axis coupling can substantially reduce the available $q$-axis voltage and shrink the feasible torque region.

\section{METHODS}
This section presents a unified framework for achieving terminal sprinting velocities in quadrupedal robots. The proposed approach consists of two components. First, a refined actuator model is developed for extreme operating conditions, explicitly capturing $d$-axis voltage coupling and magnetic saturation to reduce sim-to-real discrepancies in high-speed regimes. Second, a reinforcement learning (RL) training framework is designed for a broad command distribution. It integrates a two-stage training scheme with ACS to progressively expand the feasible velocity space while constraining command variation, thereby balancing performance and training stability as the system approaches actuator limits.

   \begin{figure}[t]
      \centering
      \includegraphics[width=\columnwidth]{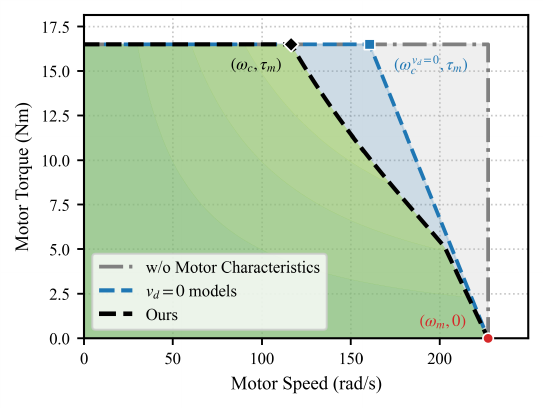}
      \caption{First-quadrant operating envelopes of a representative joint actuator under different actuator modeling paradigms.}
      \label{motor_boundary_comparison}
   \end{figure}

\subsection{Refined Actuator Modeling with \texorpdfstring{$d$}{d}-axis Voltage Coupling and Magnetic Saturation}

To ensure dynamical consistency between simulation and hardware at terminal velocities, we adopt a high-fidelity actuator model. The actuator output is strictly bounded by the MOR under high-speed, high-torque conditions. In the synchronous $dq$ reference frame (i.e., rotor-aligned), the $q$-axis current is the torque-producing component, while the $d$-axis is aligned with the rotor magnetic flux. An $i_d = 0$ control strategy is adopted, where $i_d$ and $i_q$ denote the $d$- and $q$-axis current components, respectively. For steady-state analysis, $i_q$ is treated as a constant, resulting in the following stator voltage equations:

\begin{figure*}[t]
   \centering
   \includegraphics[width=\textwidth]{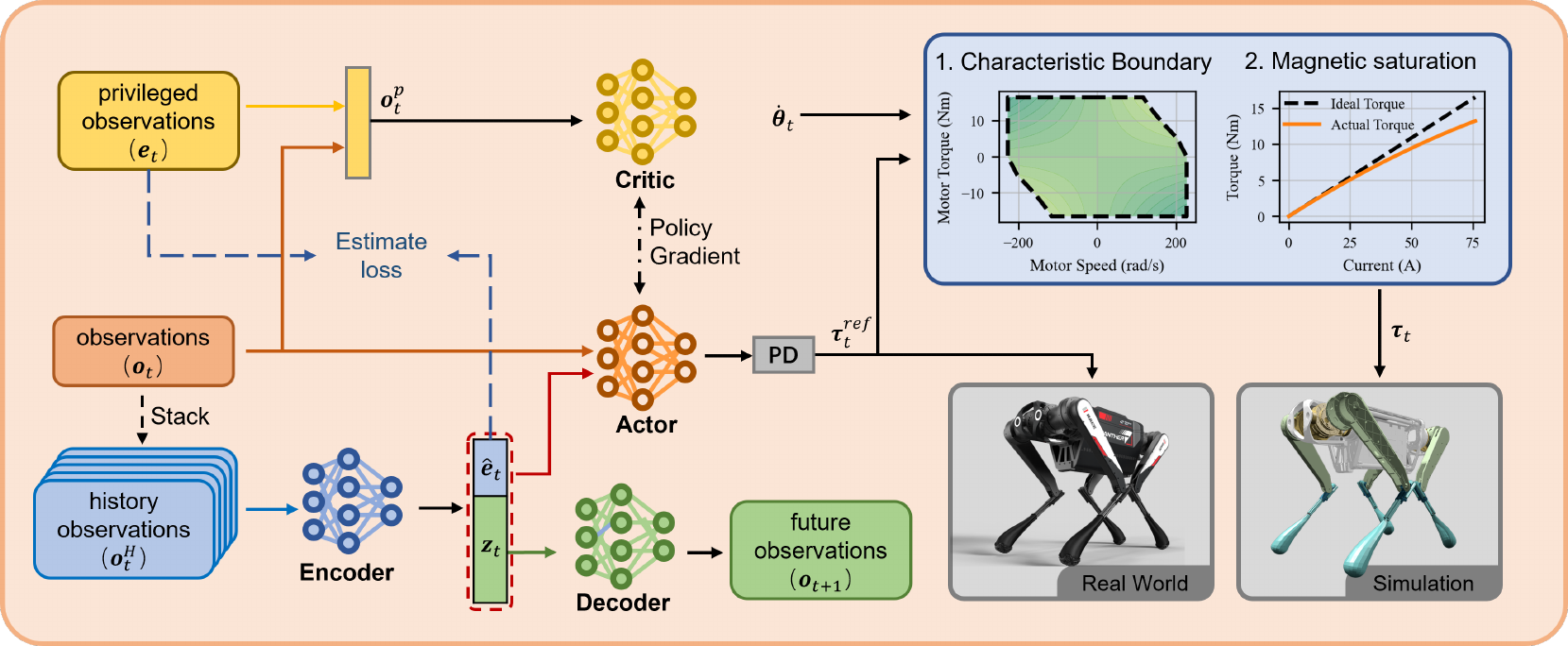}
   \caption{Training pipeline for high-speed locomotion. In simulation, policy actions are mapped to joint torques and constrained by actuator limits, including torque–speed envelope constraints and magnetic saturation. During hardware deployment, the commanded torques are directly tracked by low-level motor controllers.}
   \label{pipeline}
\end{figure*}

\begin{equation}
\label{voltage equation}
\begin{cases}
v_d &= -N \omega_{\mathrm{motor}} L_q i_q \\
v_q &= R_s i_q + N \omega_{\mathrm{motor}} \psi_m \\
\tau &= \frac{3}{2} N \psi_m i_q
\end{cases}
\end{equation}
where $v_d$ and $v_q$ denote the $d$- and $q$-axis stator voltages, $\omega_{\mathrm{motor}}$ is the motor angular velocity, $N$ is the number of pole pairs, $L_q$ is the $q$-axis inductance, $R_s$ is the stator resistance, $\psi_m$ is the permanent magnet flux linkage, and $\tau$ is the motor output torque.

Since $i_d=0$ eliminates the commanded $d$-axis current, conventional simplified actuator models often further assume that the $d$-axis voltage is negligible, i.e., $v_d=0$. Under this assumption, only the $q$-axis voltage is considered in the DC-bus voltage limit. We introduce a voltage utilization factor $\gamma=0.95$ as a conservative margin, yielding

\begin{equation}
\label{vd0 voltage constraint}
|v_q| \leq \frac{\gamma V_{\text{bus}}}{\sqrt{3}},
\end{equation}
where $V_{\text{bus}}$ denotes the DC bus voltage.

Substituting Eq.~\eqref{voltage equation} into Eq.~\eqref{vd0 voltage constraint} yields closed-form expressions for the no-load terminal speed $\omega_m$ and the critical speed $\omega_c^{v_d=0}$:

\begin{equation}
\label{vd=0 speed}
\begin{cases}
\omega_m &= \frac{\gamma V_{\text{bus}}}{\sqrt{3} N \psi_m} \\
\omega_c^{v_d=0} &= \frac{R_s}{N \psi_m} \left( \frac{\gamma V_{\text{bus}}}{\sqrt{3} R_s} - \frac{2\tau_m}{3N \psi_m} \right)
\end{cases}
\end{equation}
Here, $\omega_m$ corresponds to the no-load condition ($i_q = 0$), while $\omega_c^{v_d=0}$ denotes the maximum achievable speed under the prescribed maximum torque $\tau_m$.

However, Eq.~\eqref{voltage equation} shows that even with $i_d=0$, the induced $d$-axis voltage is not zero when the motor rotates under load. At high speeds, this component becomes non-negligible due to back-EMF and cross-axis coupling. It consumes part of the DC-bus voltage margin, reduces the voltage available to the torque-producing $q$-axis component, and consequently lowers the critical speed. Therefore, both $v_d$ and $v_q$ are retained in the voltage constraint:

\begin{equation}
\label{voltage constraint}
v_d^2+v_q^2 \leq (\frac{\gamma V_{bus}}{\sqrt{3}})^2
\end{equation}
Substituting Eq.~\eqref{voltage equation} into Eq.~\eqref{voltage constraint} yields a quadratic inequality in $\omega_{\mathrm{motor}}$ under a given $\tau$. Solving this constraint provides the maximum admissible angular velocity, from which the refined critical speed $\omega_c$ is obtained as:

\begin{equation}
\label{vc}
\omega_c = \frac{-b + \sqrt{b^2 - 4ac}}{2a}
\end{equation}
where the coefficients $a, b$, and $c$ are defined by the motor parameters and constraints:

\begin{equation}
\label{Parameter definition}
\begin{cases}
a = \dfrac{4}{9\psi_m^2} L_q^2 \tau^2_m + N^2 \psi_m^2 \\[8pt]
b = \dfrac{4}{3} R_s \tau_m \\[8pt]
c = \dfrac{4\tau_m^2 R_s^2}{9N^2\psi_m^2} - \dfrac{1}{3} \gamma^2 V^2_{\text{bus}}
\end{cases}
\end{equation}
This curvilinear boundary directly defines the motor torque saturation limits within the physics engine.

As shown in Fig.~\ref{motor_boundary_comparison}, the proposed model yields a tighter feasible torque–speed region in the high-speed regime compared to the conventional linear approximation. In particular, the linear model overestimates the admissible torque at high angular velocities, whereas the proposed formulation captures the reduction in available torque caused by $d$-axis voltage consumption. This difference leads to more restrictive torque constraints in the high-speed region.

Furthermore, the model accounts for the reduction in torque constant induced by magnetic saturation \cite{taherzadeh_torque_2021}. Unlike conventional actuator models that assume a linear torque–current relationship \cite{hwangbo_learning_2019}, the proposed approach introduces a quadratic correction fitted to experimental data to estimate the effective output torque:

\begin{equation}
\label{saturation}
\tau_{\text{output}} = \tau_c \cdot \left(1 - \alpha \frac{|\tau_c|}{\tau_m}\right)
\end{equation}
where $\tau_c$ is the commanded torque, and $\alpha$ is fitted offline from measured torque--current data.

\subsection{Reinforcement Learning Framework and Two-stage Curriculum}

To achieve stable high-speed control over a wide command range, we construct an RL pipeline with an adaptive command update mechanism.

\subsubsection{Network Architecture}

We employ an asymmetric actor–critic architecture as the algorithmic backbone \cite{nahrendra_dreamwaq_2023}, as illustrated in Fig.~\ref{pipeline}. A variational autoencoder (VAE) estimator infers a latent variable $\mathbf{z}_t \in \mathbb{R}^{16}$ and a structured estimation vector $\hat{\mathbf{e}}_t \in \mathbb{R}^8$ from historical observations. Specifically, the estimation targets include body linear velocity $\mathbf{v}_t$, base height $\mathbf{h}_t$, and foot contact states $\mathbf{c}_t$, providing physically interpretable state information and enabling rapid policy responses under highly dynamic conditions.

\begin{itemize}
\item Actor input: Comprises proprioceptive observations ($\mathbf{o}_t \in \mathbb{R}^{45}$) and estimated states. Proprioceptive observations consist of angular velocity $\boldsymbol{\omega}_t$, gravity vector $\mathbf{g}_t$, commands $\mathbf{cmd}_t$, joint positions $\boldsymbol{\theta}_t$, joint velocities $\dot{\boldsymbol{\theta}}_t$, and the previous action $\mathbf{a}_{t-1}$.
\item Critic input ($\mathbf{o}^p_t \in \mathbb{R}^{52}$): The critic receives asymmetric inputs. In addition to the actor observations, it incorporates ground-truth states $\mathbf{e}_t$ provided by the simulator.
\end{itemize}

\subsubsection{Two-stage Curriculum}

For command ranges spanning from standstill to over $10~\mathrm{m/s}$, conventional linear curriculum or uniform sampling can result in inefficient exploration in early training and policy instability in later stages due to increased nonlinear dynamics.

Accordingly, we adopt grid curriculum \cite{margolis_rapid_2022} to systematically explore the feasible locomotion envelope. Empirically, the stable locomotion envelope approximately satisfies the coupling constraint between linear velocity $v$ and angular velocity $\boldsymbol{\omega}$, bounded by the friction coefficient $\mu$, i.e., $v\omega \approx \mu g$. Based on this observation, the linear velocity range is set to $[-2, 12]~\mathrm{m/s}$ and the angular velocity range to $[-1, 1]~\mathrm{rad/s}$. Furthermore, a two-stage curriculum is designed to improve training stability and convergence:

\begin{itemize}
\item \textbf{Stage I (Fundamental Stage):} The target velocity range is incrementally expanded from $[-1, 1]~\mathrm{m/s}$ to $[-2, 6]~\mathrm{m/s}$, facilitating rapid convergence and the emergence of stable locomotion gaits.
\item \textbf{Stage II (Extreme Performance Stage):} The command range is further extended to $[-2, 12]~\mathrm{m/s}$. Additional regularization terms with adaptive weighting are introduced to account for high-speed dynamics, mitigating oscillatory behavior near actuator saturation. The reward functions follow \cite{lee_learning_2020,rudin_learning_2022,nahrendra_dreamwaq_2023}, with detailed parameters provided in Table~\ref{table_reward_final_v5}.
\end{itemize}

\begin{table}[htbp]
\caption{Reward Function Elements and Multi-Stage Weight Schedules}
\label{table_reward_final_v5}
\centering
\small
\setlength{\tabcolsep}{3pt} 
\begin{tabularx}{\columnwidth}{@{}l X cc@{}}
\toprule
\textbf{Reward Term} & \textbf{Equation} ($r_i$) & \multicolumn{2}{c}{\textbf{Weight}} \\
\midrule
\multicolumn{4}{l}{\textit{Part A: Constant Weights}} \\
\midrule
Lin. vel. track. & 
\makebox[0pt][l]{$\exp\{-(\mathbf{v}_{xy}^{cmd}-\mathbf{v}_{xy})^2\}$}\hspace{2.2cm}
& \multicolumn{2}{c}{$1.0$} \\
Ang. vel. track.   & $\exp\{-(\omega_{z}^{cmd}-\omega_{z})^2\}$             & \multicolumn{2}{c}{$0.5$} \\
Lin. vel. ($z$)    & $v_z^2$                                               & \multicolumn{2}{c}{$-2.0$} \\
Joint power        & $|\boldsymbol{\tau} \cdot \dot{\boldsymbol{\theta}}|$        & \multicolumn{2}{c}{$-1.0 \times 10^{-6}$} \\
Action rate        & $(\mathbf{a}_t - \mathbf{a}_{t-1})^2$                     & \multicolumn{2}{c}{$-0.01$} \\
Smoothness         & $(\mathbf{a}_t - 2\mathbf{a}_{t-1} + \mathbf{a}_{t-2})^2$ & \multicolumn{2}{c}{$-0.005$} \\ 
Joint pos. lim.    & $(\boldsymbol{\theta} - \boldsymbol{\theta}_{lim})^2$      & \multicolumn{2}{c}{$-10.0$} \\
\midrule
\multicolumn{2}{l}{\textit{Part B: Multi-Stage Weight Schedules}} & \textbf{Stage 1} & \textbf{Stage 2} \\
\midrule
Ang. vel. ($xy$)   & $\omega_x^2 + \omega_y^2$                     & $-0.02$ & $-0.5$ \\
Orientation        & $\|\mathbf{g}_{xy}\|^2$                       & $-5$    & $-100$ \\
Base height        & $(\mathbf{h} - h_{target})^2$                             & $-10$   & $-20$ \\
Joint acc.         & $\ddot{\boldsymbol{\theta}}^2$            & $-2.5 \cdot 10^{-7}$ & $-1.0 \cdot 10^{-7}$ \\
Abad pos.          & $\boldsymbol{\theta}_{abad}^2$                           & $-1.0$  & {\scriptsize N/A} \\
Collision          & $n_{\text{col}}$                               & $-1.0$  & {\scriptsize N/A} \\
Foot slip          & $|\mathbf{v}_{xy}^\text{foot}| \cdot \mathbf{c}$    & {\scriptsize N/A} & $-0.01$ \\
Yaw rate           & $\omega_z^2$                                   & {\scriptsize N/A} & $-0.5$ \\
Foot clearance     & $(z_{f} - z_{tg})^2$                     & $-0.2$  & {\scriptsize N/A} \\
Abad mirror        & $(\boldsymbol{\theta}^{L}_{abad} + \boldsymbol{\theta}^{R}_{abad})^2$    & {\scriptsize N/A} & $-0.5$ \\
\bottomrule
\multicolumn{4}{p{\columnwidth}}{\footnotesize \textbf{Note:} Weights in Part A are constant. In Part B, N/A indicates the term is inactive.}
\end{tabularx}
\end{table}

\subsubsection{Adaptive Command Scheduling}

In high-speed locomotion, the command update mechanism critically affects policy stability and performance. Conventional random command sampling induces a mismatch between commanded velocity and system state, which is exacerbated under extreme dynamics: large tracking weights drive aggressive pursuit of infeasible commands, causing instability, whereas small weights yield conservative policies that remain stationary to avoid penalties. Fixed or stochastic acceleration constraints partially mitigate this mismatch but rely on heuristic tuning, tend to overfit specific profiles, and fail to capture nonlinear, state-dependent acceleration limits induced by actuator saturation, contact dynamics, and inertia; thus, they are inadequate for exploiting physical limits in high-speed regimes.

To address this limitations, we propose an ACS strategy based on the instantaneous body velocity $v_{\text{curr}}$
, constraining the next command within a local neighborhood:
\begin{equation}
\label{acs}
v_{\text{cmd}, t+1} \in [v_{\text{curr}, t} - \Delta v,\; v_{\text{curr}, t} + \Delta v].
\end{equation}

This formulation implicitly regulates the rate of change of the command, aligning updates with instantaneous motion capability and ensuring smooth evolution of the tracking objective during training. Compared with random sampling and heuristic constraints, ACS reduces tracking error, avoids aggressive or trivial policies, stabilizes optimization, and improves learning reliability in high-speed regimes. In experiments, $\Delta v=1$ balances responsiveness and stability, enabling smooth transitions without excessive tracking lag.

   \begin{figure}[t]
      \centering
      \includegraphics[width=\columnwidth]{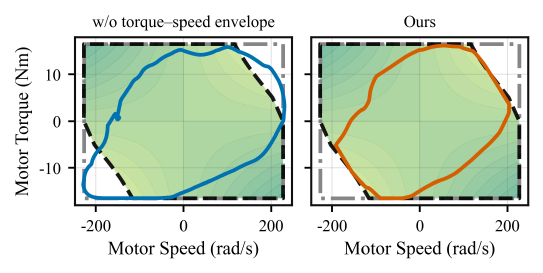}
      \caption{Comparison between the baseline without torque–speed envelope constraints and the proposed method in their respective training environments, where both policies operate at the limits of actuator performance.}
      \label{ideal_motor}
   \end{figure}

   \begin{figure}[htbp]
      \centering
      \includegraphics[width=\columnwidth]{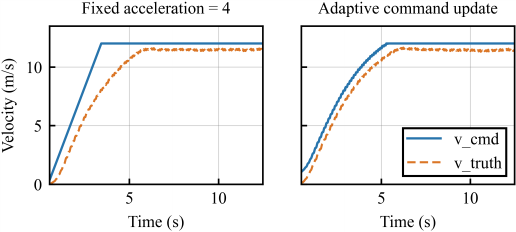}
      \caption{Visualization of locomotion patterns under diverse command scheduling modes enabled by the proposed ACS. The method generalizes across different command update paradigms while preserving acceleration profiles consistent with the robot dynamics.}
      \label{cmd_comparsion}
   \end{figure}

\section{EXPERIMENTS}

To evaluate the effectiveness of our proposed framework, high-speed locomotion experiments are conducted in simulation as well as in indoor and outdoor real-world settings.

\subsection{Experimental Setup}

\textbf{Simulation Training:}
Training is performed in parallel across 4096 environments using IsaacGym on a single RTX 4090 GPU. A set of DR is applied on both observations and physical parameters.
In contrast to position-based control schemes commonly used in prior work \cite{liu_discrete-time_2025}, direct torque control is employed to model the dynamics of high-stiffness passive spring joints. Soft constraints are introduced near workspace boundaries to emulate contact with mechanical joint limits. Specifically, once the joint position enters a predefined threshold region near the limit, the effective stiffness is rapidly increased, approximating rigid contact behavior.

\textbf{Hardware Deployment:}
The control framework is deployed on the custom-developed quadrupedal platform, BP2. The robot has a total mass of $36.5 \mathrm{kg}$ and operates with a $60 \mathrm{V}$ DC bus. The high-level policy runs onboard at $50 \mathrm{Hz}$, while low-level motor control operates at $1000 \mathrm{Hz}$. For high-speed locomotion, PD gains for the hip joints (thigh and calf) are set to $K_p = 160$, $K_d = 4.9$, while the abduction joints use $K_p = 60$, $K_d = 4.9$. To reduce sensitivity to observation noise, a 10~Hz low-pass filter is applied to proprioceptive observations, while a 3~Hz filter is applied to state estimation. These estimated states provide auxiliary motion feedback rather than high-bandwidth stabilization signals, and the lower cutoff helps suppress state-estimation noise and vibration-induced disturbances during high-speed hardware deployment.

\begin{table}[!t]
\caption{Comparison of Velocity and Stability Metrics}
\label{table_performance_comparison}
\centering
\small
\setlength{\tabcolsep}{3pt}
\begin{tabularx}{0.5\textwidth}{@{}>{\raggedright\arraybackslash}X>{\centering\arraybackslash}p{0.11\textwidth}>{\centering\arraybackslash}p{0.15\textwidth}@{}}
\toprule
Method & Max Velocity (m/s) & Relative Stability ($J_{\mathrm{norm}}$) \\
\midrule
Baseline & 9.4 & 0.876 \\
w/o torque–speed envelope & 11.0 & 1.212 \\
Simplified DC & 11.7 & 1.070 \\
Fixed acceleration scheduling & --- & --- \\
Uniform command sampling & 11.0 & 1.452 \\
\textbf{Ours} & \textbf{12.4} & \textbf{1.000} \\
\bottomrule
\multicolumn{3}{p{0.48\textwidth}}{\footnotesize \textbf{Note:} The Fixed acceleration scheduling method induces training instability and catastrophic collapse, thereby preventing evaluation.}
\end{tabularx}
\end{table}

\begin{figure*}[!t]
  \centering
  \includegraphics[width=\textwidth]{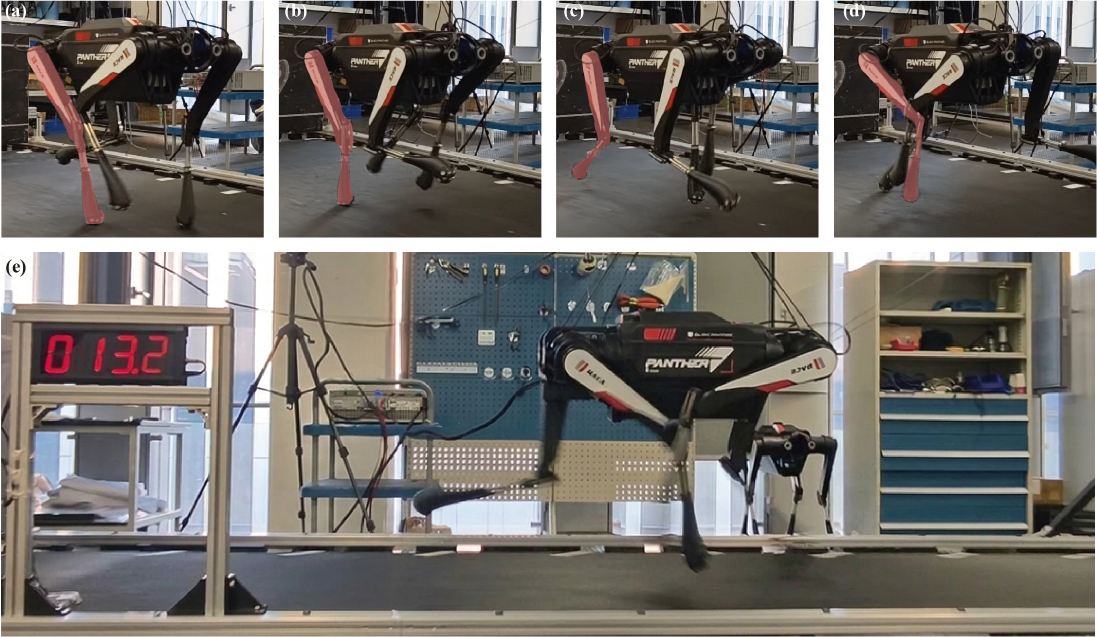}
  \caption{Experimental setup and representative snapshots from indoor locomotion experiments. (a)–(d) depict one complete gait cycle. (e) Snapshot at the instant when the robot reaches its maximum velocity.}
  \label{indoors}
\end{figure*}

To validate each component of the proposed framework, we conduct same-platform ablations on BP2. All methods use the same reward terms, command range, training budget, and evaluation protocol as Ours, except for the replaced components listed below:

\begin{itemize}
\item \textbf{Baseline:} replaces both the refined actuator model and ACS with a simplified square torque--speed envelope and uniform command sampling.
\item \textbf{w/o torque--speed envelope:} replaces the refined actuator model with a simplified square torque--speed envelope.
\item \textbf{Simplified DC model:} replaces the refined actuator model with a conventional DC-equivalent model that omits the $d$-axis voltage component.
\item \textbf{Fixed acceleration scheduling:} replaces ACS with command updates sampled under a fixed acceleration range of $[2,4]~\mathrm{m/s^2}$.
\item \textbf{Uniform command sampling:} replaces ACS with uniform command sampling over the full command range.
\end{itemize}

\subsection{Simulation Experiments}

Sim-to-sim validation is conducted in MuJoCo under realistic dynamic constraints. As direct measurement of body linear acceleration on hardware is non-trivial, evaluation adopts a standardized acceleration–steady-state–deceleration protocol with a constant acceleration bound of $4~\mathrm{m/s^2}$, enabling systematic assessment of transient response and near-limit stability while improving sim-to-real fidelity.

Due to the absence of a waist joint, maintaining a level base is critical for high-speed locomotion. Stability is therefore evaluated via base angular velocity:

\begin{equation}
   \label{stable}
   J = \int_{0}^{T} (\omega_{\text{pitch}}^2 + \omega_{\text{roll}}^2) dt
\end{equation}
where $\omega_{\text{pitch}}$ and $\omega_{\text{roll}}$ are measured in $\mathrm{rad/s}$. A normalized metric $J_{\mathrm{norm}} = J / J_{\text{Ours}}$ is used for consistent cross-method comparison.

As summarized in Table~\ref{table_performance_comparison}, the proposed method achieves the highest maximum velocity of $12.4~\mathrm{m/s}$, improving over the baseline by $31.9\%$. Although the baseline has a lower $J_{\mathrm{norm}}$ of $0.876$, it is prone to forward pitching during high-acceleration phases, limiting its achievable speed. In contrast, removing torque--speed envelope constraints or replacing ACS with uniform command sampling substantially increases $J_{\mathrm{norm}}$, indicating degraded base smoothness near the actuator limits. These results show that accurate actuator-envelope modeling and ACS are both necessary for extending high-speed locomotion while maintaining stable motion.

These discrepancies arise from inaccurate actuator limit modeling: neglecting torque–speed constraints yields unrealizable envelopes that the policy exploits during training, whereas hardware saturation enforces clipping, causing power deficits, oscillatory dynamics, and amplified pitch oscillations near terminal velocity (Fig.~\ref{ideal_motor}). This underscores the necessity of accurate actuator modeling for sim-to-real consistency.

Moreover, fixed acceleration constraints induce overfitting to specific command profiles, leading to failure under the full evaluation protocol. In contrast, ACS promotes state-dependent acceleration consistent with system dynamics—low during startup for smooth transitions, higher at mid-speed for efficiency, and reduced near terminal velocity for stability (Fig.~\ref{cmd_comparsion}).

\subsection{Indoor Experiments}

In indoor experiments, high-speed locomotion tests are conducted on a custom treadmill, with belt velocity measured via the transmission system. For safety, lateral tethers were used only to restrict excessive yaw and lateral drift during acceleration, while the overhead tether served only as fall protection. They did not provide propulsion, posture control, or body-weight support, and were not engaged when the robot reached the peak speed.
As shown in Fig.~\ref{indoors}, the results show that the proposed method achieves a stable sprinting speed of up to $13.2 \mathrm{m/s}$, surpassing all previously reported speeds.

To further analyze gait characteristics at high speed, a representative gait cycle at peak velocity is selected for Fig.~\ref{indoors}(a)–(d). The figure illustrates the motion phases of the right hind leg within one cycle. At touchdown, the foot impacts the ground and joint elasticity absorbs impact energy, which is stored as elastic potential energy. During the stance phase, the leg swings backward while the stored energy is progressively released, generating propulsive force through ground reaction. After liftoff, during the swing phase and subsequent contralateral touchdown, the leg swings forward rapidly to achieve a large stride length in preparation for the next contact. This cyclic process enables sustained propulsion and coordination at high speed, highlighting the role of joint elasticity in high-velocity locomotion.

\subsection{Outdoor Experiments}

Outdoor experiments are conducted on a concrete surface to evaluate real-world performance and sim-to-real consistency. Body linear velocity is measured at $10~\mathrm{Hz}$ using a high-precision RTK system. The experimental protocol follows the same \emph{acceleration–steady-state–deceleration} profile as in simulation.

\begin{figure}[!t]
\centering
\includegraphics[width=\columnwidth]{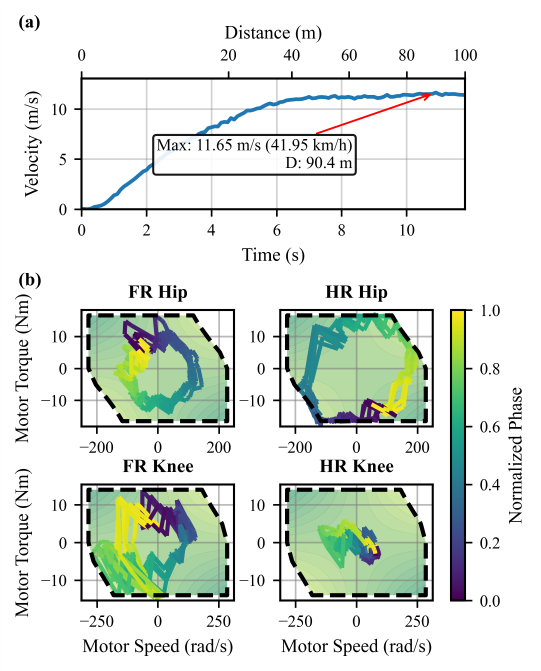}
\caption{Outdoor locomotion experiment results. (a) Time series of body velocity measured by the RTK system. (b) Corresponding torque–speed profiles of representative actuators, illustrating operation near actuator limits and close agreement with simulated behaviors.}
\label{outdoors}
\end{figure}

As shown in Fig.~\ref{outdoors}(a), the robot completes a $100~\mathrm{m}$ sprint in $11.67~\mathrm{s}$, reaching a peak speed of $11.65~\mathrm{m/s}$, with smooth acceleration, stable high-speed locomotion, and controlled deceleration. Fig.~\ref{outdoors}(b) further provide actuator-level analysis, showing that the hip joints—particularly the hind-leg motors—operate close to their physical limits during high-speed motion. The close agreement between simulated and measured actuator behavior indicates that the proposed actuator model accurately captures the underlying hardware characteristics, even near operational boundaries.

As summarized in Table~\ref{table_speed_comparison_outdoor}, BP2 achieves the highest reported outdoor speed among quadrupedal robots while maintaining a small sim-to-real gap. In contrast, prior methods typically exhibit larger discrepancies between simulation and hardware performance. This comparison, together with the actuator-level consistency observed in Fig.~\ref{outdoors}, highlights the necessity and advantage of precise actuator boundary modeling for achieving both high-speed performance and reliable sim-to-real transfer.

\begin{table}[t]
\caption{Comparison of Reported Maximum Speeds in Simulation and Outdoor Hardware}
\label{table_speed_comparison_outdoor}
\centering
\small
\setlength{\tabcolsep}{4pt} % 稍微增大一点
\begin{tabularx}{\columnwidth}{@{}>{\raggedright\arraybackslash}X >{\centering\arraybackslash}X >{\centering\arraybackslash}X@{}}
\toprule
Method & Max Sim Speed (m/s) & Max Outdoor Speed (m/s) \\
\midrule
WildCat~\cite{noauthor_introducing_nodate} & --- & 8.89 \\
RLvRL~\cite{margolis_rapid_2022} & 5.7 & 3.9 \\
HACL~\cite{mishra_hacl_2025} & 6.7 & 4.1 \\
Hound~\cite{shin_reinforcement_2025} & 6.5 & 5.9 \\
Spot~\cite{miller_high-performance_2025} & --- & 5.2 \\
\textbf{Ours} & \textbf{12.4} & \textbf{11.65} \\
\bottomrule
\multicolumn{3}{p{\columnwidth}}{\footnotesize \textbf{Note:} Reported speeds are taken directly from the corresponding publications. ``---'' indicates that a directly comparable maximum simulation speed is not explicitly reported.}
\end{tabularx}
\end{table}

\section{DISCUSSION}

The experimental results indicate that actuator limits are an important factor determining the upper bound of high-speed locomotion performance in quadrupedal robots. Simulation studies without accurate torque–speed envelope constraints show that relaxing hip joint limits improves the operating conditions of the knee joints, revealing strong inter-joint coupling in load distribution. These findings suggest that explicitly modeling and potentially expanding the MOR, for example via flux-weakening control, may further improve high-speed performance and energy efficiency without requiring hardware modifications.

However, the results also suggest that current control strategies rely heavily on hip actuator output. To address this limitation, two directions for future work are outlined: First, offline trajectory optimization (TO) can be employed with joint load balancing as an objective to derive sprinting gaits with more uniform power distribution. Incorporating such trajectories as priors for reinforcement learning \cite{peng_amp_2021} may guide the policy to utilize actuator limits more effectively while maintaining balanced power allocation. Second, from a mechanical design perspective, introducing additional degrees of freedom, such as an active waist joint, may further improve actuation utilization. Combined with numerical optimization of mass distribution and link parameters, this approach may reduce the load on individual joints and extend the system’s locomotion capability.

\section{CONCLUSION}

This work addresses terminal-velocity locomotion in quadrupedal robots with a control framework tailored for extreme dynamic regimes. A high-fidelity actuator model captures $d$-axis voltage coupling and magnetic saturation, reducing high-speed sim-to-real discrepancies. Building on this model, a reinforcement learning framework combines a two-stage curriculum and ACS to improve training stability across broad command distributions and mitigate command-induced instability.

Experiments on the BP2 platform demonstrate stable high-speed locomotion near actuator limits and a peak outdoor speed of $11.65~\mathrm{m/s}$, among the fastest reported for quadrupedal robots, confirming the framework's effectiveness and practical deployability.

These results underscore the importance of accurate actuator modeling for unlocking extreme locomotion performance and provide a foundation for future high-speed legged-robot research.

%%%%%%%%%%%%%%%%%%%%%%%%%%%%%%%%%%%%%%%%%%%%%%%%%%%%%%%%%%%%%%%%%%%%%%%%%%%%%%%%

%%%%%%%%%%%%%%%%%%%%%%%%%%%%%%%%%%%%%%%%%%%%%%%%%%%%%%%%%%%%%%%%%%%%%%%%%%%%%%%%

%%%%%%%%%%%%%%%%%%%%%%%%%%%%%%%%%%%%%%%%%%%%%%%%%%%%%%%%%%%%%%%%%%%%%%%%%%%%%%%%

\bibliographystyle{IEEEtran} % use IEEEtran.bst style
\bibliography{IEEEabrv, ./BP2CONTROLLER.bib}

\addtolength{\textheight}{-12cm}   % This command serves to balance the column lengths
                                  % sure that you do not shorten the textheight too Wmuch.

\end{document}